\pdfoutput=1

\documentclass[11pt]{article}
\usepackage{afterpage}
\usepackage{graphicx} 
\usepackage{arabtex}
\usepackage{utf8}
\setcode{utf8} 
\usepackage{float}
\usepackage{amsmath}
\usepackage{makecell}
\usepackage{xspace}
\usepackage{subcaption}
\usepackage{tcolorbox}
\usepackage{booktabs}
\usepackage{adjustbox}

\usepackage{times}
\usepackage{latexsym}
\usepackage{todonotes}
\usepackage{multirow}
\usepackage[shortlabels]{enumitem}
\usepackage[T1]{fontenc}
\usepackage{microtype}
\usepackage{inconsolata}
\usepackage{xcolor}
\usepackage{textcomp} 
\usepackage{upquote} 
\usepackage{acl2023}

\definecolor{c1}{HTML}{4daf4a}
\definecolor{c2}{HTML}{377eb8}
\definecolor{c3}{HTML}{984ea3}
\definecolor{c4}{HTML}{a65628}
\definecolor{c5}{HTML}{ffff33}
\definecolor{c6}{HTML}{ff7f00}
\definecolor{c7}{HTML}{e41a1c}

\newcolumntype{P}[1]{>{\centering\arraybackslash}p{#1}}

\title{
Curriculum Learning and Pseudo-Labeling Improve the Generalization of Multi-Label Arabic Dialect Identification Models}

\author{
  Ali Mekky\thanks{\quad Equal contribution.} \quad
  Mohamed El Zeftawy\footnotemark[1] \quad
  Lara Hassan\footnotemark[1]  \quad
  Amr Keleg \quad
  Preslav Nakov\\
  Mohamed bin Zayed University of Artificial Intelligence, Abu Dhabi, UAE \\
  \texttt{\{firstname.lastname\}@mbzuai.ac.ae}
}

\begin{document}
\maketitle
\setcounter{footnote}{0}
\begin{abstract}

Being modeled as a single-label classification task for a long time, recent work has argued that Arabic Dialect Identification (ADI) should be framed as a multi-label classification task. However, ADI remains constrained by the availability of single-label datasets, with no large-scale multi-label resources available for training. By analyzing models trained on single-label ADI data, we show that the main difficulty in repurposing such datasets for Multi-Label Arabic Dialect Identification (MLADI) lies in the selection of negative samples, as many sentences treated as negative could be acceptable in multiple dialects. To address these issues, we construct a multi-label dataset by generating automatic multi-label annotations using GPT-4o and binary dialect acceptability classifiers, with aggregation guided by the Arabic Level of Dialectness (ALDi). Afterward, we train a BERT-based multi-label classifier using curriculum learning strategies aligned with dialectal complexity and label cardinality. On the MLADI leaderboard, our best-performing \textsc{LahjatBERT} model achieves a macro F1 of 0.69, compared to 0.55 for the strongest previously reported system. Code and data are available at \url{https://mohamedalaa9.github.io/lahjatbert/}.
\end{abstract}

\section{Introduction}
Arabic has a wide range of diverse dialects spoken across the Arab World. While Modern Standard Arabic (MSA) is used in official communication, education, and media, everyday conversations typically happen in local dialects \cite{habash2010introduction}. Dialects vary between countries, and can even vary within same-country cities and local communities, creating a complex linguistic landscape
\cite{10.1162/COLI_a_00169, althobaiti2020automaticarabicdialectidentification}.

Arabic Dialect Identification (ADI) is the task that aims to identify the dialect of a sentence. ADI has been modeled as a single-label classification problem, with systems trained and evaluated on resources such as \texttt{AOC} \cite{zaidan-callison-burch-2011-arabic}, \texttt{MADAR} \cite{bouamor-etal-2019-madar}, \texttt{QADI} \cite{abdelali-etal-2021-qadi}, and \texttt{NADI} benchmarks \cite{abdul-mageed-etal-2020-nadi, abdul-mageed-etal-2021-nadi, abdul-mageed-etal-2022-nadi, abdul-mageed-etal-2023-nadi}. However, as illustrated in \autoref{tab:dialect-examples}, a single utterance may simultaneously sound natural to speakers from multiple countries, making it inherently multi-dialectal and difficult for traditional single-label classification methods to produce accurate results \cite{keleg-magdy-2023-arabic,olsen-etal-2023-arabic}.
Even moderately long sentences can be acceptable in multiple dialects \cite{keleg-etal-2025-revisiting}.

\begin{table}[t!]
    \centering
    \adjustbox{max width=\linewidth}{
    \begin{tabular}{@{}p{3.5cm}p{5.5cm}@{}}
        \textbf{Dialects} & \textbf{Sentence} \\ 
        \midrule
        Jordan, Palestine
        & \makebox[5.5cm][r]{\RL{اتفقنا نظل جنب بعض ع الحلوه والمره}} \newline
          (We agreed to stick by each other through thick and thin.) \\ 
        \midrule
        Algeria, Egypt, Jordan, Palestine, Sudan, Syria, Tunisia, Yemen
        & \makebox[5.5cm][r]{\RL{احسبو حسابي معاكم}} \newline
          (Make sure to count me in with you.) \\ 
        \midrule
        Algeria, Palestine, Sudan, Yemen
        & \makebox[5.5cm][r]{\RL{بسمتك يا زين تسوي ألف بسمه}} \newline
          (Your smile, oh beautiful one, is worth a thousand other smiles.) \\ 
        \midrule
        All Arabic Dialects
        & \makebox[5.5cm][r]{\RL{اللهم الجنه لمن ذهبت أرواحهم إليك}} \newline
          (O Allah, grant Paradise to those whose souls have returned to You.) \\ 
        \bottomrule
    \end{tabular}
    }
    \caption{
    Examples illustrating dialect overlap in Arabic, sampled from the manually annotated \texttt{NADI2024} Subtask~1 development set. \textbf{Note:} the dataset has labels for eight country-level dialects only.
    }
    \label{tab:dialect-examples}
\end{table}

To better capture this phenomenon, the field has begun transitioning to \emph{Multi-Label Arabic Dialect Identification} (MLADI), which allows an utterance to be tagged as acceptable in multiple dialects.\footnote{Multi-label Dialect Identification is being explored for other languages such as French and Spanish \cite{bernier-colborne-etal-2023-dialect,zampieri-etal-2024-language,lopetegui-etal-2025-common}.} 
This transition has been driven by the evolution of the \texttt{NADI} shared tasks, a recurring benchmark series for Arabic Dialect Identification. 
While earlier editions of NADI (2020–2023) framed ADI as a single-label classification problem, assigning each sentence to a single country-level dialect \cite{abdul-mageed-etal-2020-nadi,abdul-mageed-etal-2021-nadi,abdul-mageed-etal-2023-nadi}, the \texttt{NADI2024} shared task introduced multi-label annotations to explicitly account for dialectal overlap \cite{abdul-mageed-etal-2024-nadi}. 
However, in \texttt{NADI2024}, multi-label annotations are provided only for the development and test sets, while the training data remains single-labeled, creating a mismatch between the nature of the task and the structure of the training data. 
As a result, MLADI introduces a weakly-supervised classification problem, where models must learn to predict multiple dialects based on datasets that provide only one ground-truth label per instance.

In this work, we examine the mismatch between the multi-label nature of Arabic dialect usage and the single-label structure of existing ADI datasets by analyzing the behavior of binary dialect-specific acceptability classifiers trained on single-label NADI datasets \cite{abdul-mageed-etal-2020-nadi,abdul-mageed-etal-2021-nadi,abdul-mageed-etal-2023-nadi}. Using the classifiers’ training dynamics \cite{swayamdipta-etal-2020-dataset}, we show that a substantial portion of samples treated as negative supervision is in fact judged acceptable by native speakers. This observation highlights the difficulty of accurately selecting negative samples when repurposing single-label, geo-located datasets for multi-label dialect identification and motivates explicit multi-label supervision.

Building on these insights, we construct a multi-label training set by aggregating pseudo-labels from two heterogeneous sources: GPT-4o and a set of 18 binary dialect acceptability classifiers. Using this dataset, we train a BERT-based multi-label classifier, which we refer to as \textsc{LahjatBERT}. We train it under three settings: without curriculum learning, with an ALDi-aware curriculum, and with a label-cardinality-based curriculum \citep{10.1145/1553374.1553380}. The two curriculum strategies progressively expose the model to sentences of increasing dialectal ambiguity, allowing it to learn from simpler instances before confronting harder ones.

Our contributions are threefold:
\begin{enumerate}[label=\arabic*., leftmargin=*, noitemsep, nolistsep]
    \item We provide an in-depth analysis of the limitations of reusing single-label ADI datasets for multi-label dialect acceptability.
    
    \item We construct a pseudo-labeled multi-label dataset for MLADI by combining the predictions of GPT-4o and binary dialect classifiers.\footnote{Following the NADI shared-task license, we release only tweet IDs and derived labels, not the underlying tweet text.}
    
    \item We introduce \textsc{LahjatBERT}, a family of multi-label BERT-based models trained on the constructed dataset, and investigate curriculum-based training variants aligned with the multi-label structure of MLADI. The best-performing variant achieves $69.04\%$ macro-F1 on the MLADI leaderboard~\cite{keleg-etal-2025-revisiting}, surpassing the top \texttt{NADI2024} system and outperforming larger multilingual and Arabic-specific language models.

\end{enumerate}

\section{MLADI Task's Setup and Previous Attempts}

\begin{figure}[tbph!]
    \centering
    \includegraphics[width=0.9\columnwidth]{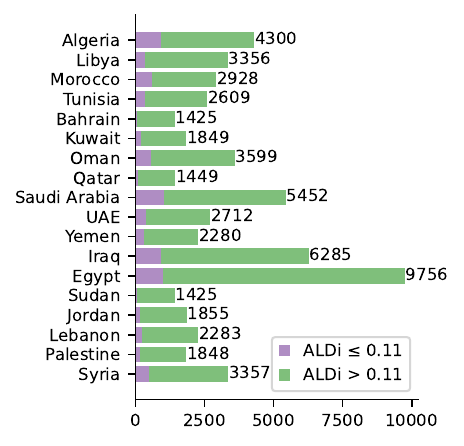}
    \caption{Number of samples in each dialect after combining the NADI~2020,~2021,~2023 datasets. Samples with automatically estimated \textit{Arabic Level of Dialectness} (ALDi; \protect\citealp{keleg-etal-2023-ALDi}) $\leq 0.11$ are expected to be in MSA. The majority of the MSA samples are expected to be acceptable in all dialects.}
    \label{fig:combined-distribution}
\end{figure}

The MLADI dataset only provides a development set of 120 samples and a test set of 1000 samples. Each sample of the development set is manually labeled by annotators from 8 countries, while the test set has 11 country-level acceptability labels. To build multi-label ADI systems, the task requires using the following three single-labeled NADI datasets: NADI~2020 \cite{abdul-mageed-etal-2020-nadi}, NADI~2021 \cite{abdul-mageed-etal-2021-nadi}, and NADI~2023 \cite{abdul-mageed-etal-2023-nadi}. All these datasets provide tweets annotated with their estimated geo-location country label covering 18 countries: Algeria, Bahrain, Egypt, Iraq, Jordan, Kuwait, Lebanon, Libya, Morocco, Oman, Palestine, Qatar, Saudi Arabia, Sudan, Syria, Tunisia, the UAE, and Yemen. Since these datasets rely on the user geo-location rather than manual linguistic annotation, posts authored by users whose geo-location differs from their country of origin can be mislabeled \cite{abdul-mageed-etal-2024-nadi}.
NADI 2023 contains 1{,}000 tweets for each of 18 Arabic country-level dialects. By contrast, NADI 2020 and NADI 2021 are more imbalanced, with dialects such as Bahraini and Qatari underrepresented relative to more frequent varieties like Egyptian and Iraqi. The combined distribution of samples across dialects is shown in \autoref{fig:combined-distribution}.

\paragraph{Previous MLADI Attempts.}
\citet{kanjirangat-etal-2024-nlp} used a nearest-neighbour approach to predict multiple labels for each sample, by encoding the training data samples and the test set into the same embedding space.
\citet{karoui-etal-2024-elyadata} tackled the weak supervision problem by applying a similarity-based label expansion strategy. Their method, SIMMT, heuristically assigns additional labels to each sample based on vocabulary similarity between dialects, followed by multi-label fine-tuning of transformer models. This approach demonstrated that moving from single-label to multi-label supervision can improve performance on the MLADI task.
Our work replaces heuristic label expansion with a pseudo-labeling framework that integrates complementary signals from multiple models to produce multi-label annotations. This yields a richer supervision signal than surface-level similarity measures.

\section{Difficulties of Using Existing Datasets for Dialect Acceptability Classification}
\label{sec:difficulties}

An intuitive idea to build multi-label ADI systems is to repurpose single-label ADI datasets for training multiple independent binary classifiers, each of which assesses the acceptability of a sentence in a specific dialect. For a specific country-level dialect (\textit{Cntry}), one might take samples geolocated to \textit{Cntry} as \textbf{positive} (i.e., \textit{acceptable}) samples for the country's classifier, and samples geolocated to other countries as \textbf{negatives}. Previous attempts have shown that this technique does not achieve the best classification performance \cite{karoui-etal-2024-elyadata,kanjirangat-etal-2024-nlp}. However, they have not analyzed the reasons for the failure of this technique, which we investigate here.
\subsection{Other Countries’ Samples Are Not Always Negative Samples}
\paragraph{Definition} For a dialect acceptability model, the negative class represents \textit{sentences with \textbf{any linguistic feature} (e.g., a morpheme or a lexical item) that is \textbf{not acceptable} in the considered dialect}. The positive class represents the remaining sentences acceptable in this dialect or in MSA.

The majority of the samples geo-located to the considered dialect are expected to be positive samples. The fact that some of these samples could also be acceptable in other dialects does not impact their categorization as positive samples, according to the aforementioned definition. To the contrary, considering samples geo-located to other dialects as negative samples is problematic. More specifically, a subset of these samples is expected to also be acceptable in the dialect considered. Consequently, this subset of negative sentences should be reassigned to the positive class. We next show how model training dynamics could guide the identification of wrongly assigned negative samples.\footnote{\citet{abdul-mageed-etal-2024-nadi} found that NADI's geo-location method has a moderate to high accuracy of ensuring a sentence's acceptability in a specific country-level dialect.}

\begin{figure*}[tb]
    \centering
    \begin{subfigure}{.25\textwidth}
      \centering
      \includegraphics[width=0.95\textwidth]{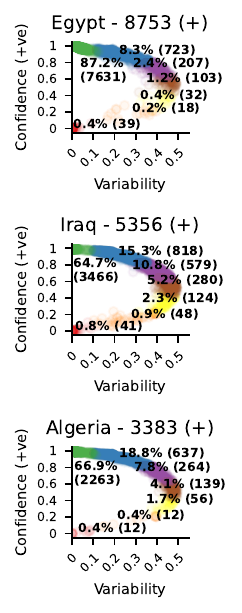}
    \end{subfigure}%
    \begin{subfigure}{.25\textwidth}
      \centering
      \includegraphics[width=0.95\textwidth]{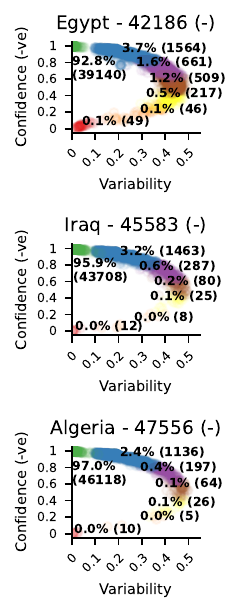}
    \end{subfigure}%
    \unskip\vrule
    \begin{subfigure}{.25\textwidth}
      \centering
      \includegraphics[width=0.95\textwidth]{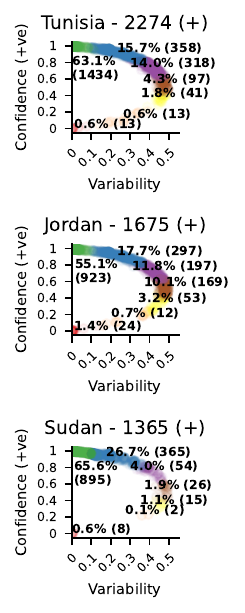}
    \end{subfigure}%
    \begin{subfigure}{.25\textwidth}
      \centering
      \includegraphics[width=0.95\textwidth]{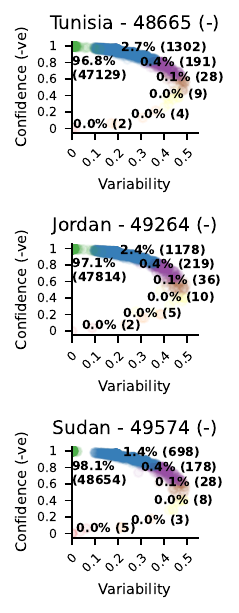}
    \end{subfigure}%
    \caption{The training dynamics for 6 binary acceptability classifiers, characterized by the mean confidence in the label across different steps/stages of the model training (y-axis), and the standard deviation of these confidence values (x-axis). Each pair shows the training dynamics' metrics for the non-MSA positive (left) and negative (right) samples of a single classifier, with the respective number of samples shown above each subplot. \textbf{Note:} Sample's correctness ranges are \fcolorbox{black}{c7}{\rule{0pt}{3pt}\rule{3pt}{0pt}}: 0 \fcolorbox{black}{c6}{\rule{0pt}{3pt}\rule{3pt}{0pt}}: ]0, 0.2[ \fcolorbox{black}{c5}{\rule{0pt}{3pt}\rule{3pt}{0pt}}: [0.2, 0.4[ \fcolorbox{black}{c4}{\rule{0pt}{3pt}\rule{3pt}{0pt}}: [0.4, 0.6[ \fcolorbox{black}{c3}{\rule{0pt}{3pt}\rule{3pt}{0pt}}: [0.6, 0.8[ \fcolorbox{black}{c2}{\rule{0pt}{3pt}\rule{3pt}{0pt}}: [0.8, 1[ \fcolorbox{black}{c1}{\rule{0pt}{3pt}\rule{3pt}{0pt}}: 1
    }
    \label{fig:training_dynamics}
\end{figure*}

\subsection{Training Dynamics and Multi-label Samples}
\citet{swayamdipta-etal-2020-dataset} used three metrics tracked during the model training process to categorize each sample's difficulty to be learned. For a training sample x\textsubscript{i} with a label y\textsubscript{i}, the confidence that the model assigns to the target y\textsubscript{i} is tracked at the end of each training epoch (or every number `\textit{N}' of training batches). The mean and standard deviation of the tracked confidence scores assigned to the sample's target label are termed the \textbf{\textit{Confidence}} and \textbf{\textit{Variability}} metrics, respectively. \textbf{\textit{Correctness}} represents the percentage of epochs (or \textit{N} batches) for which the model assigns a higher probability to the sample's target label than all the other labels.
Based on these metrics, the training dataset is split into three main categories---\textit{(1)~easy to learn} samples: high-\textit{Confidence} and low-\textit{Variability}, \textit{(2)~ambiguous} samples: moderate-\textit{Confidence} and high-\textit{Variability}, and \textit{(3)~hard to learn}: low-\textit{Confidence} and low-\textit{Variability}.

\paragraph{Intuition} A dialect acceptability classifier would struggle to learn negative samples that are also acceptable in the considered dialect, as these samples should belong to the positive class. These samples are expected to be \textit{ambiguous} or even \textit{hard to learn}.

\paragraph{Methodology} We train 18 dialect acceptability classifiers, one for each country represented in the NADI datasets. For each country's classifier, samples of the NADI datasets geo-located to this country, and MSA samples (ones with ALDi < 0.11) are considered as positive (acceptable) samples, with the remaining samples considered as negative (unacceptable) samples. Confidence scores are tracked every 300 steps ($\approx \frac{1}{5}$th~of~an~epoch) for 5 epochs. However, the first epoch's confidence scores are ignored, as the model's training dynamics could be unstable during the early learning stages \cite{swayamdipta-etal-2020-dataset}.

\paragraph{Findings}\autoref{fig:training_dynamics} shows the training dynamics for six different acceptability classifiers, for the positive (non-MSA) samples and the negative samples. First, a few samples (<50) for each set have a correctness value of 0. Moreover, the larger the imbalance between the number of positive and negative samples becomes, the smaller the percentage of negative samples with non-perfect correctness is. For instance, 7.2\% (n=3,046) of Egypt's negative samples have non-perfect correctness scores compared to only 1.9\% (n=920) for Sudan. These negative samples with non-perfect correctness scores are potentially wrongly assigned samples.

\subsection{Usability of Training Dynamics in Flagging Wrongly Assigned Samples}
To understand the effectiveness of training dynamics in identifying wrongly assigned samples, we engaged native speakers to manually evaluate the acceptability of sentences with varying correctness scores. Specifically, we recruited one annotator from the following countries: Egypt~(EG), Iraq~(IQ), Algeria~(DZ), Tunisia~(TN), Jordan~(JO), and Sudan~(SD).\footnote{Notably, the six recruited annotators previously participated in the annotation of NADI~2024's evaluation sets.}

For each country, we sample 10 examples from each of the seven correctness ranges: \textit{0, ]0,~0.2[, [0.2,~0.4[, [0.4,~0.6], [0.6,~0.8[, [0.8,~1[, 1}. In total, 140 samples are annotated for each dialect (70 of which are positive and the remaining 70 are negative), unless one of the bins had fewer than 10 samples. The annotators follow the same guidelines of NADI~2024's first~subtask~\cite{abdul-mageed-etal-2024-nadi}, whereby they answer: \textit{\textbf{Is it possible that the tweet is authored by someone who speaks one of your country’s dialects?}
Options:~Yes, Not~Sure/Maybe, or No}.

\paragraph{Results} Inspecting the number of acceptable samples in each bin in \autoref{tab:dynamics_check}, a large number of negative samples with \textit{correctness} $<1$ are rated as acceptable, indicating the effectiveness of using non-perfect correctness scores to flag wrongly assigned negative samples. Conversely, a majority of the positive samples are rated as acceptable for the different correctness bins, even for correctness scores that are less than 0.6, which could be attributed to the class imbalance toward negative samples.

\begin{table}[!tbhp]
    \centering
    \small
    \setlength{\tabcolsep}{1pt} 

        \begin{tabular}{lcc|ccccc|c}
            \multirow{2}{*}{} & \multirow{2}{*}{\textbf{Label}} & \multicolumn{7}{c}{\textbf{Correctness}} \\
            & & \textbf{0} & \textbf{]0,0.2[} & \textbf{[0.2,0.4[} & \textbf{[0.4,0.6[} & \textbf{[0.6,0.8[} & \textbf{[0.8,1[} & \textbf{1} \\
            & & \fcolorbox{black}{c7}{\rule{0pt}{3pt}\rule{3pt}{0pt}} & \fcolorbox{black}{c6}{\rule{0pt}{3pt}\rule{3pt}{0pt}} & \fcolorbox{black}{c5}{\rule{0pt}{3pt}\rule{3pt}{0pt}} & \fcolorbox{black}{c4}{\rule{0pt}{3pt}\rule{3pt}{0pt}} & \fcolorbox{black}{c3}{\rule{0pt}{3pt}\rule{3pt}{0pt}} & \fcolorbox{black}{c2}{\rule{0pt}{3pt}\rule{3pt}{0pt}} & \fcolorbox{black}{c1}{\rule{0pt}{3pt}\rule{3pt}{0pt}} \\ 
            \midrule
            \multirow{2}{*}{\rotatebox{90}{\textbf{EG}}} & \textcolor{blue}{+ve} & \textcolor{red}{2} & \textcolor{blue}{6} & \textcolor{blue}{6} & \textcolor{blue}{6} & \textcolor{blue}{7} & \textcolor{blue}{9} & \textcolor{blue}{10} \\ 
             & \textcolor{red}{-ve} & \textcolor{blue}{10} & \textcolor{blue}{10} & \textcolor{blue}{10} & \textcolor{blue}{9} & \textcolor{blue}{8} & \textcolor{red}{4} & \textcolor{red}{2} \\ 
            \midrule
            \multirow{2}{*}{\rotatebox{90}{\textbf{IQ}}} & \textcolor{blue}{+ve} & \textcolor{red}{4} & \textcolor{red}{2} & \textcolor{red}{1} & \textcolor{blue}{6} & \textcolor{red}{4} & \textcolor{black}{5} & \textcolor{blue}{7} \\ 
             & \textcolor{red}{-ve} & \textcolor{blue}{9} & \textcolor{blue}{6/8} & \textcolor{blue}{10} & \textcolor{blue}{9} & \textcolor{blue}{7} & \textcolor{black}{5} & \textcolor{red}{2} \\ 
            \midrule
            \multirow{2}{*}{\rotatebox{90}{\textbf{DZ}}} & \textcolor{blue}{+ve} & \textcolor{red}{3} & \textcolor{red}{4} & \textcolor{black}{5} & \textcolor{black}{5} & \textcolor{blue}{8} & \textcolor{blue}{7} & \textcolor{blue}{9} \\ 
             & \textcolor{red}{-ve} & \textcolor{blue}{9} & \textcolor{blue}{5/5} & \textcolor{blue}{8} & \textcolor{blue}{9} & \textcolor{blue}{9} & \textcolor{blue}{8} & \textcolor{red}{4} \\ 
            \midrule
            \multirow{2}{*}{\textbf{\rotatebox{90}{\textbf{TN}}}} & \textcolor{blue}{+ve} & \textcolor{red}{1} & \textcolor{red}{3} & \textcolor{red}{1} & \textcolor{red}{0} & \textcolor{red}{1} & \textcolor{black}{5} & \textcolor{blue}{7} \\ 
             & \textcolor{red}{-ve} & \textcolor{blue}{2/2} & \textcolor{blue}{4/4} & \textcolor{blue}{8/9} & \textcolor{blue}{7} & \textcolor{black}{5} & \textcolor{red}{4} & \textcolor{red}{1} \\ 
            \midrule
            \multirow{2}{*}{\rotatebox{90}{\textbf{JO}}} & \textcolor{blue}{+ve} & \textcolor{red}{3} & \textcolor{blue}{7} & \textcolor{blue}{6} & \textcolor{black}{5} & \textcolor{blue}{8} & \textcolor{blue}{10} & \textcolor{blue}{9} \\ 
             & \textcolor{red}{-ve} & \textcolor{blue}{2/2} & \textcolor{blue}{5/5} & \textcolor{blue}{10} & \textcolor{blue}{10} & \textcolor{blue}{9} & \textcolor{blue}{9} & \textcolor{red}{2} \\ 
            \midrule
            \multirow{2}{*}{\rotatebox{90}{\textbf{SD}}} & \textcolor{blue}{+ve} & \textcolor{blue}{5/8} & \textcolor{black}{1/2} & \textcolor{blue}{6} & \textcolor{blue}{7} & \textcolor{blue}{10} & \textcolor{blue}{8} & \textcolor{blue}{10} \\ 
             & \textcolor{red}{-ve} & \textcolor{blue}{5/5} & \textcolor{blue}{3/3} & \textcolor{blue}{8/8} & \textcolor{blue}{7} & \textcolor{blue}{8} & \textcolor{blue}{10} & \textcolor{red}{1} \\ 
            \bottomrule
        \end{tabular}%
    \caption{The number of acceptable samples for the different correctness bins. For each row, each bin contains 10 total samples, except for a few bins marked by the \textit{acceptable/total} format. \textbf{Note:} \textcolor{blue}{Values marked in blue} indicate that >50\% of the bin's samples are acceptable, while \textcolor{red}{values marked in red} indicate that <50\% of the bin's samples are acceptable. The high-correctness bins of the positive class are expected to have a large number of acceptable samples. Conversely, the high-correctness bins of the negative class are expected to have a small number of acceptable samples, which is not the case. Country codes follow the ISO~3166-1 alpha-2 standard.}
    \label{tab:dynamics_check}
\end{table}

\paragraph{Moving Forward} Our analysis shows that a large proportion of negative samples with non-perfect correctness scores should be reassigned to the positive class. However, manual annotation is still required to assess these automatically flagged samples. Moreover, the class imbalance seems to have an impact on the model's dynamics. Future work could consider using a \textit{human and model in the loop} setup \cite{vidgen-etal-2021-learning}. More specifically, the initial model's training dynamics are used to automatically identify ambiguous samples, which are then manually reassigned to the correct class. Afterward, another model is trained from scratch on the dataset after reassignment, with the new model's dynamics used to automatically identify new ambiguous samples.

\section{Multi-Label ADI Dataset Creation}
\label{sec:dataset}

It seems inevitable that building multi-label ADI models requires the presence of multi-label ADI training datasets, especially to reduce the number of false negatives for each dialect (i.e., samples wrongly assumed unacceptable in that dialect). However, building large enough multi-label ADI datasets is expensive, since annotating just 1,120 samples by speakers of 9 countries could cost as much as \$1,700 \cite{abdul-mageed-etal-2024-nadi}. Hence, we propose two pseudo-labeling methods for building multi-label ADI datasets:

\paragraph{(1) Binary Dialect Classifiers.} We again build 18 independent acceptability classifiers, one per country-level dialect $dia$, each predicting whether a sentence $x$ is acceptable in dialect $dia$. These classifiers are trained only on the balanced NADI~2023 dataset to avoid skewed pseudo-labels. For a sentence $x_{i}$, we generate 18 acceptability pseudo-labels using the 18 classifiers as:\\ $\hat{\mathbf{y}}^{\mathrm{BIN}}_{x_i} =
\left( Accept_{dia_{1}}(x_i), \dots, Accept_{dia_{18}}(x_i) \right).$

Here, we notably rely on the Arabic Level of Dialectness \cite[ALDi;][]{keleg-etal-2023-ALDi} score as a global signal characterizing the degree of dialectness of each sentence ($a_i~=~\mathrm{ALDi}(x_i) \in [0,1]$). \citet{keleg-etal-2025-revisiting} showed that ALDi moderately correlates with the number of dialects in which a sentence is acceptable. The higher the ALDi score of a sentence, the less the number of dialects in which the sentence is acceptable.

For the \textbf{positive} samples of $dia$'s acceptability classifier $Accept_{dia}(x_{i})$, 
we consider all sentences from the NADI datasets that are geolocated to $dia$, in addition to MSA sentences with $a_i < 0.11$, which are broadly acceptable across dialects, irrespective of their geolocation.
Since treating all sentences geolocated to
other regions as negative examples leads to systematic errors (as shown in Section~\ref{sec:difficulties}), we prioritize the precision of selecting true \textbf{negative samples}. To this end, we only select sentences that are (1)~geolocated to \emph{non-neighbouring} dialect regions, (2)~with high dialectness ($a_i > 0.77$), where linguistic overlap with dialect $dia$ is less likely.\footnote{See Appendices~\ref{app:ALDi-thresholds},\ref{app:non-neighbouring} and \ref{app:diagnostics} for more implementation details.}

\paragraph{(2) GPT-Based Multi-Label Annotation.}
For each sentence, we obtain multi-label annotations from GPT-4o by prompting the model
to independently assess the acceptability of the sentence in each of the 18 dialects.
This yields a binary relevance vector
$
\hat{\mathbf{y}}^{\mathrm{GPT}}_i \in \{0,1\}^{18}.
$
The full prompting template is provided in Appendix~\ref{appendix:prompt}.

\subsection{Quality of Supervision Signals}
\label{sec:supervision-quality}

To evaluate the two labeling methods, we use the development set of the \texttt{NADI2024} shared task
\cite{abdul-mageed-etal-2024-nadi}, which has 120 sentences  annotated for 8
country-level dialects. This provides a useful benchmark for evaluating the performance of the two methods, as shown in \autoref{tab:method_dev_results}.

\begin{table}[tb]
\centering
\small
\setlength{\tabcolsep}{1pt} 
\begin{tabular}{p{4.5cm}cccc}
\textbf{Pseudo-Labeling Method} & \textbf{P\textsubscript{macro}} & \textbf{R\textsubscript{macro}} & \textbf{F1\textsubscript{macro}} & \textbf{Acc.} \\
\midrule

(1) Binary Dialect Classifiers & \textbf{78.2} & 51.4 & 60.4 & 76.2 \\
(2) GPT-4o Pseudo-Labels & 73.5 & \textbf{66.3} & 67.8 & 77.4 \\
\midrule
Hybrid ALDi-Based Labels (\S\ref{sec:aggregation}) & 77.5 & 62.8 & \textbf{68.5} & \textbf{79.3} \\
\bottomrule
\end{tabular}
\caption{Macro-averaged Precision, Recall, F1, and Accuracy of the two methods in addition to the aggregation method (\S\ref{sec:aggregation}) on the \texttt{NADI2024}
development set. 
}
\label{tab:method_dev_results}
\end{table}

Our conservative way of selecting the samples for the binary classifiers results in a high macro precision of 78.2 and a moderate macro recall of 51.4. In contrast, GPT-4o achieves substantially higher recall and a better overall F1 score. This hints that GPT-4o is more willing to assign multiple dialect labels, particularly in cases where dialectal overlap is present.

\subsection{Pseudo-Labels Aggregation}
\label{sec:aggregation}

To better understand the results in \autoref{tab:method_dev_results}, we analyze
performance across ALDi ranges in \autoref{tab:ALDi-binned-supervision}. For
MSA samples ($[0,0.11)$), both GPT-4o and the Binary Classifiers achieve perfect
precision. For highly dialectal samples ($[0.77,1]$), the Binary Classifiers attain
higher precision, indicating more reliable negative supervision when strong
dialectal cues are present.

In contrast, GPT-4o consistently achieves higher recall and F1 in the intermediate
ALDi ranges ($[0.11,0.77]$), where dialectal overlap is more common. This behavior
is consistent with the conservative construction of the Binary Classifiers, whose
negative samples are restricted to highly dialectal cases ($a_i>0.77$), improving
precision at the extremes but limiting coverage in the mid-range.

\begin{table}[hptb]
\centering
\small
\setlength{\tabcolsep}{1.25pt} 
\begin{tabular}{clccc}
\textbf{ALDi Bin} & \textbf{Supervision Source} & \textbf{P\textsubscript{macro}} & \textbf{R\textsubscript{macro}} & \textbf{F1\textsubscript{macro}} \\
\midrule
\multirow{1}{*}{$[0, 0.11)$}      & (1) Binary Classifiers  & 100.00 & \textbf{94.64} & \textbf{97.12} \\
(n=7) & (2) GPT-4o  & 100.00 & 90.48 & 94.84 \\
\midrule
\multirow{1}{*}{$[0.11, 0.44)$}     & (1) Binary Classifiers  & 76.99 & 46.81 & 55.99 \\
(n=16)& (2) GPT-4o  & \textbf{84.72} & \textbf{61.36} & \textbf{69.72} \\
\midrule
\multirow{1}{*}{$[0.44, 0.77)$}     & (1) Binary Classifiers  & 63.28 & 36.72 & 44.38 \\
(n=48)& (2) GPT-4o  & \textbf{63.86} & \textbf{56.94} & \textbf{57.15} \\
\midrule
\multirow{1}{*}{$[0.77, 1.0]$}    & (1) Binary Classifiers  & \textbf{77.60} & 51.38 & 59.03 \\
(n=49)& (2) GPT-4o  & 62.36 & \textbf{68.14} & \textbf{62.62} \\
\bottomrule
\end{tabular}
\caption{Performance of the two pseudo-labeling methods on subsets of the \texttt{NADI2024} development set.}
\label{tab:ALDi-binned-supervision}
\end{table}

Consequently, we aggregate the predictions of the two pseudo-labeling methods by
using the Binary Classifiers at the ALDi extremes ($a_i<0.11$ or $a_i>0.77$) and
GPT-4o in the intermediate range ($0.11\le a_i\le 0.77$). This aggregation combines
the strengths of both methods and achieves the best macro F1 and accuracy on the
evaluation set (last row of \autoref{tab:method_dev_results}), reflecting an
improved precision--recall trade-off.

\begin{figure}[t]
    \centering
    \includegraphics[width=0.9\linewidth]{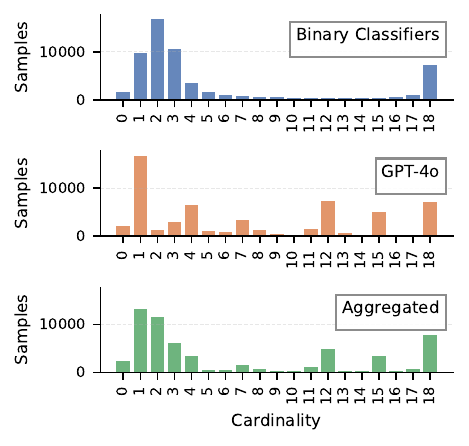}
    \caption{
    Number of samples for each label cardinality according to the three pseudo-labeling methods.
    }
    \label{fig:cardinality-distribution}
\end{figure}

\section{Multi-Label Dialect Classification Models}
\label{sec:multilabel-modeling}

Using the pseudo-labeled multi-label ADI dataset described in Section~\ref{sec:dataset}, we fine-tune MARBERT~\cite{abdul2021arbert} as a multi-label classifier over the 18 Arabic dialects (country-level ADI). We refer to the resulting family of models fine-tuned under different settings, as \textsc{LahjatBERT}.
We train using binary cross-entropy with logits loss, treating each label independently. During fine-tuning, we freeze the bottom 8 transformer layers of MARBERT and update only the top 4 layers and the classification head. We do not fully fine-tune all MARBERT layers since fine-tuning only a fraction
of the final layers recovers most downstream effectiveness \cite{lee2019elsadofreezinglayers}. Also, we discard zero-cardinality samples (i.e., instances for which the pseudo-labeling step assigns no dialect), since
keeping them would treat the samples as negative for all 18 dialects and inject systematic noise. We report the complete training and inference hyperparameters in \autoref{app:hyperparameters}.

An analysis of the dataset (\autoref{fig:cardinality-distribution})
 reveals a strong skew in the cardinality distribution: most samples contain only one or two active dialects. This bias might encourage the model
to predict low-cardinality outputs, substantially reducing recall for sentences acceptable in multiple dialects.
To mitigate the dataset’s inherent low-cardinality bias,
we adopt two strategies for curriculum learning (CL). In both strategies, the complexity of the training examples is increased in a controlled manner during the model training process, to achieve better model generalization \cite{10.1145/1553374.1553380}. The exact ordering of both the cardinality buckets and the ALDi buckets is motivated in \autoref{appendix:curriculum motivation}.

\subsection{Cardinality-Based Curriculum Learning}
In this strategy, the samples' label cardinality is used as a proxy for their respective complexities. For a sentence \(x_i\) with a pseudo-labeled target vector \(\mathbf{y}_i \in \{0,1\}^{18}\), the label cardinality 
$c(x_i)~=~\|\mathbf{y}_i\|_0$
represents the number of dialects in which the sentence is acceptable. This strategy is proposed to mitigate the skewness of the training dataset samples toward lower cardinalities.

We partition the training set into cardinality buckets \(B_c\), where \(B_c\) contains samples with label cardinality \(c\).
We then define a curriculum ordering over these buckets using the mean loss of each cardinality.
Let \(\pi(e)\) denote the cardinality index of the bucket selected at curriculum stage \(e\).
At stage \(e\), the model is trained on (1) all examples from \(B_{\pi(e)}\), and (2) an equal number of randomly sampled examples from each previously introduced bucket \(B_{\pi(1)}, \dots, B_{\pi(e-1)}\).

\begin{figure}[!bthp]
    \centering
    \includegraphics[width=0.9\linewidth]{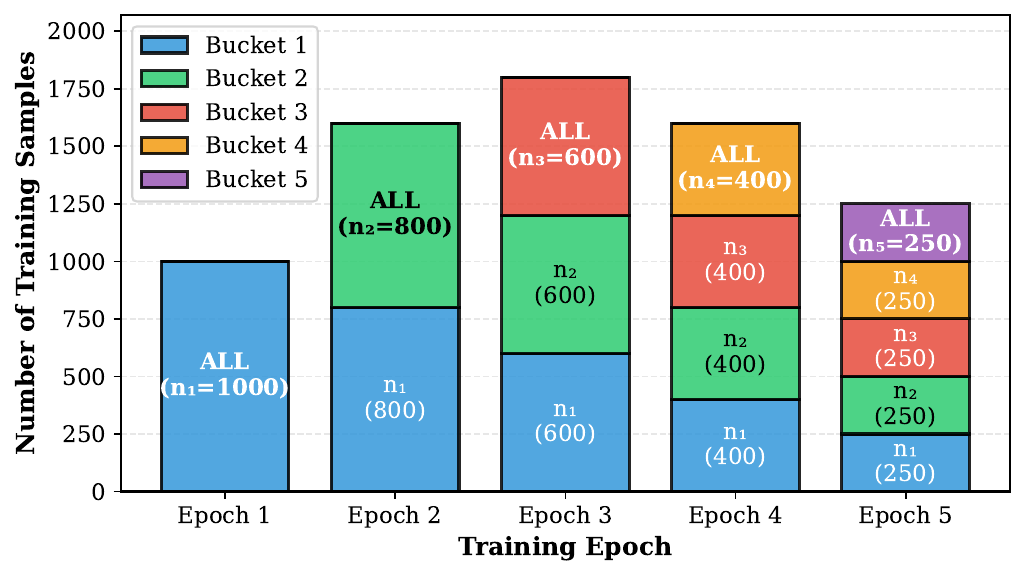}
    \caption{
    Illustration of the cardinality-based curriculum schedule, showing the progressive introduction of higher difficulty cardinality samples. 
    The numerical values are illustrative and do not reflect the actual dataset.
    }
    \label{fig:cardinality_curriculum}
\end{figure}

The model first trains exclusively on the bucket introduced at the first curriculum stage. In the next stage, the bucket selected next is introduced together with an equal number of samples drawn from each previously introduced bucket. This progression continues until the full cardinality range in the dataset is incorporated. The overall training schedule is illustrated in \autoref{fig:cardinality_curriculum}, which visualizes how buckets are gradually introduced while maintaining balanced exposure across earlier stages.

\subsection{ALDi-Based Curriculum Learning}
ALDi provides another intuitive proxy for the samples' complexities. Samples with intermediate ALDi scores are expected to be harder than MSA samples with low scores and high-score samples that show clear cues of a specific dialect.
 
We divide the continuous ALDi range into four contiguous intervals:
$I_1~= [0,~0.11), I_2~= [0.11,~0.44), 
I_3~= [0.44,~0.77), I_4~= [0.77,~1].$
For each interval $I_k$, we construct a bucket $B_k$ that contains all training examples whose ALDi score falls in that range:
$B_k = \{ x_i \mid a(x_i) \in I_k \}$; where $a(x_i)~\in~[0,1]$ denotes sentence $x_i$'s ALDi score.

We define a curriculum ordering over the ALDi buckets based on average training loss.
Let $\pi(e)$ denote the index of the ALDi bucket selected at epoch $e$, with buckets ordered from lowest to highest loss.
At epoch $e$, the model is trained on (1) all examples from the current bucket $B_{\pi(e)}$, and (2) a random subset of examples from each previously introduced bucket $B_{\pi(1)}, \dots, B_{\pi(e-1)}$, following the same sampling strategy as the cardinality-based CL.

\section{Results}
\label{sec:results}

We evaluate our multi-label dataset construction and training strategies on the
\texttt{NADI2024} development set, which provides multi-label annotations for
\textbf{8 country-level dialects}. Since our training and supervision span \textbf{18 dialects},
we report metrics only on the overlapping label set.

\subsection{Multi-Label Model Performance}
\label{sec:multilabel-performance}

We evaluate how a single multi-label classifier can learn from different
supervision signals by training the same model architecture on each dataset without
curriculum learning. This setup isolates the effect of the supervision labels
themselves.

As shown in \autoref{tab:nadi-results} (Block~II), the model trained on GPT-4o labels
achieves higher macro Recall but lower Precision, indicating a tendency to
over-predict dialect labels. In contrast, training on the hybrid labels yields
higher Precision and Accuracy while maintaining comparable macro F1. This behavior
is consistent with the design of the hybrid labels, which combines the
conservative behavior of the binary classifiers at ALDi extremes with the richer
GPT-based labels in the ALDi mid-range.

\begin{table}[H]
\centering
\small
\setlength{\tabcolsep}{1pt} 
\begin{tabular}{p{4cm}cccc}
\textbf{Model} & \textbf{P\textsubscript{macro}} & \textbf{R\textsubscript{macro}} & \textbf{F1\textsubscript{macro}} & \textbf{Acc.} \\
\midrule
\multicolumn{5}{c}{\textbf{(I) Baseline}} \\
\midrule
NADI 2024 Baseline &71.2 & 30.9 & 39.7 & 69.3 \\
\midrule
\multicolumn{5}{c}{\textbf{(II) Multi-Label Model Performance}} \\
\midrule
Trained on GPT-4o labels & 69.6 & \textbf{65.7} & 66.2 & 75.5 \\
Trained on hybrid labels & \textbf{73.7} & 63.7 & \textbf{67.4} & \textbf{77.5} \\
\midrule
\multicolumn{5}{c}{\textbf{(III) Curriculum Learning on Hybrid Dataset}} \\
\midrule
Hybrid + Cardinality-Based CL & 69.0 & \textbf{80.6} & \textbf{72.7} & 77.5 \\
Hybrid + ALDi-Based CL & \textbf{71.4} & 71.3 & 70.3 & \textbf{78.2} \\
\bottomrule
\end{tabular}
\caption{Macro-averaged Precision, Recall, F1, and Accuracy on the \texttt{NADI2024}
development set. Results are grouped into (i) multi-label
model performance without CL, and (ii) the effect of CL.}
\label{tab:nadi-results}
\end{table}

\subsection{Impact of Curriculum Learning}
\label{sec:curriculum-results}

We next examine the effect of curriculum learning by comparing curriculum-based
training with the baseline setting without a curriculum. As shown in
\autoref{tab:nadi-results} (Block~III), both curriculum learning strategies improve
macro F1 relative to training without curriculum learning, while consistently increasing recall and
reducing precision.

In our setup, the curriculum order is derived from a loss-based criterion
(Appendix~§\ref{appendix:curriculum motivation}), where examples
introduced at the latest stages tend to be more ambiguous, corresponding to
intermediate label cardinalities and intermediate ALDi scores.

One possible interpretation of the observed recall increase is that exposure at
the latest curriculum stages to higher-loss, more ambiguous examples encourages
the model to activate a larger set of labels per instance. These examples are
characterized by intermediate ALDi scores and multiple valid dialect labels per
instance. This broader label coverage may help recover more relevant dialect
labels, leading to higher recall. At the same time, predicting more labels per
instance can make the model less conservative, which is reflected in the
accompanying reduction in precision.

This interaction also helps explain why curriculum learning is most effective when
applied to the hybrid supervision. Since the hybrid annotations exhibit higher
precision than GPT-4o labels alone, they help limit the precision loss associated
with increased recall.

\subsection{Generalization on the MLADI Test Set}

We evaluate whether the improvements on the development set generalize
to the MLADI test set. \autoref{tab:mladi_test_results} reports the performance
of the three \textsc{LahjatBERT} variants compared with the NADI~2024 baseline
and previously reported systems.

The NADI 2024 baseline is a single-label dialect identification model, but it is converted into a multi-label predictor at test time. 
For each sentence, it computes a softmax distribution over the 18 dialect classes, then selects the most likely dialects until their cumulative probability reaches a fixed value (we use the Top-90\% setting, i.e., $P{=}0.9$), and returns those dialects as the predicted label set \cite{abdul-mageed-etal-2024-nadi}.

All \textsc{LahjatBERT} variants outperform the shared-task baseline and prior
approaches in terms of macro F1, indicating that the gains obtained from the
constructed supervision and training strategies extend beyond the development
setting. Among the \textsc{LahjatBERT} variants, the ALDi-based curriculum achieves the
highest macro F1, while the cardinality-based curriculum yields the highest recall,
mirroring the precision-recall trade-offs observed on the development set.\footnote{We analyze the models' predictions in the Appendix (§\ref{sec:predictions_contrast}).}

\begin{table}[H]
\centering
\small
\setlength{\tabcolsep}{1pt} 
\begin{tabular}{lcccc}
\textbf{Model} & \textbf{P\textsubscript{macro}} & \textbf{R\textsubscript{macro}} & \textbf{F1\textsubscript{macro}} & \textbf{Acc.} \\
\midrule
\textsc{LahjatBERT} (no curriculum) & \textbf{69.0} & 69.7 & 68.0 & \textbf{79.1} \\
\textsc{LahjatBERT} + Cardinality CL & 59.3 & \textbf{81.0} & 66.6 & 73.9 \\
\textsc{LahjatBERT} + ALDi CL & 65.0 & 76.4 & \textbf{69.0} & 78.2 \\
\midrule
Aya-32B~\cite{dang2024ayaexpansecombiningresearch} & 49.5 & 64.5 & 54.5 & 65.6 \\
Elyadata~\cite{karoui-etal-2024-elyadata} & 50.2 & 56.9 & 52.4 & 67.0 \\
NADI~2024 Baseline & 64.8 & 39.9 & 47.0 & 72.3 \\
\bottomrule
\end{tabular}
\caption{Macro-averaged Precision, Recall, F1, and Accuracy on the MLADI test set~\cite{keleg-etal-2025-revisiting}, which contains 1,000 sentences annotated for 11 dialects.}
\label{tab:mladi_test_results}
\end{table}

\section{Conclusion and Future Work}

This work examines the limitations of reusing single-label Arabic dialect identification (ADI) datasets for multi-label ADI (MLADI). We specifically highlight that the difficulty of building binary dialect acceptability classifiers lies in the selection of negative samples. While training dynamics can help in automatically flagging wrongly-assigned negative samples, manual verification is still required to assess the acceptability of these flagged samples.
Using single-label ADI data, we construct a pseudo-labeled multi-label dataset by aggregating predictions from GPT-4o and binary dialect acceptability classifiers, and introduce \textsc{LahjatBERT}, a family of BERT-based multi-label models that outperform existing MLADI systems.

Future work could incorporate targeted human annotation for samples identified as ambiguous by the analysis, enabling iterative refinement of multi-label supervision.
Finally, our CL approach substantially improves recall. Future work could examine whether similar gains extend to other multi-label tasks beyond dialect identification.

\section*{Limitations}
Our evaluation is limited by the label coverage of available benchmarks. Although our training and supervision signals span 18 dialects, the \texttt{NADI2024} development set provides multi-label annotations for only 8 country-level dialects, so development metrics are computed only on the overlapping label subset. Likewise, the test set covers only 11 dialects, restricting conclusions about generalization to the full 18-dialect label space.

A second limitation originates from the fact that NADI relies on user geo-location as a proxy for dialect rather than direct linguistic annotation, which can be noisy when posting location and actual dialect do not align.

\section*{Ethics and Broader Impact}

This work uses anonymized Arabic tweets from the publicly released NADI shared-task datasets, collected via the Twitter API under the platform’s terms. Following NADI licensing restrictions, we release only tweet IDs and our automatically generated multi-label annotations (not the underlying tweet text), along with code and trained models to enable reproducibility. The research is methodological, and while dialect identification can support beneficial language technologies and sociolinguistic analysis, it also carries risks of misuse (e.g., profiling or surveillance), which we explicitly do not endorse and is outside the scope of this work.

\bibliography{custom}
\bibliographystyle{acl_natbib}

\appendix
\setcounter{table}{0}
\setcounter{figure}{0}
\renewcommand{\thetable}{\Alph{section}\arabic{table}}
\renewcommand{\thefigure}{\Alph{section}\arabic{figure}}

\section{Experimental Details}
\label{app:hyperparameters}
We provide the full training and inference hyperparameters for the multi-label setup described in \autoref{sec:multilabel-modeling}.
We fine-tune a BERT-based model for multi-label country-level Arabic dialect identification with 18 binary labels, where the classification head outputs one logit per country. Training minimizes binary cross-entropy with logits, which implements a combination of sigmoid and binary cross-entropy and treats each label independently. We split the data into 90\% training and 10\% validation with a fixed random seed of 42. During fine-tuning, we freeze the bottom 8 transformer layers of MARBERT and update only the top 4 layers and the classification head. We set both the hidden-state and attention dropout probabilities to 0.3. We train for 3 epochs with a batch size of 24 for both training and evaluation, evaluating once per epoch. The best checkpoint is selected based on validation micro F1. At inference time, we apply a sigmoid to the logits and use a threshold of 0.3 to obtain binary label assignments. We chose this value by maximizing validation micro F1 on the held-out split, the default threshold of 0.5 yields lower validation macro F1, so we use 0.3 for all reported results.

\section{ALDi Threshold Selection}
\label{app:ALDi-thresholds}

We adopt two ALDi thresholds, $a_i = 0.11$ and $a_i = 0.77$, to distinguish sentences with
minimal dialectal evidence from those with strong dialectal evidence. These thresholds
are grounded in the annotation scheme and definition of the Arabic Level of Dialectness
(ALDi) score introduced by \citet{keleg-etal-2023-ALDi}.

\paragraph{ALDi Annotation Scheme.}
Following the ALDi annotation protocol described by \citet{keleg-etal-2023-ALDi},
each sentence is annotated by three native Arabic speakers. Annotators assign one of four
ordinal labels reflecting the degree of dialectness: \textit{MSA} (0), \textit{Little}
($\tfrac{1}{3}$), \textit{Mixed} ($\tfrac{2}{3}$), and \textit{Most} (1). The ALDi score
for a sentence is defined as the mean of the three annotations. Consequently, ALDi values
lie in $[0,1]$ and take discrete steps of $\tfrac{1}{9}$.

\paragraph{Derivation of the Thresholds.}
Under this definition, the smallest non-zero ALDi value is
\[
\frac{0 + 0 + \tfrac{1}{3}}{3} = \tfrac{1}{9} \approx 0.11,
\]
which corresponds to the weakest possible evidence of dialectal content, where two
annotators label the sentence as MSA and one assigns a \textit{Little} dialect label.
Conversely, an ALDi score of
\[
\frac{1 + 1 + \tfrac{1}{3}}{3} = \tfrac{7}{9} \approx 0.77
\]
corresponds to strong dialectal evidence, where two annotators assign the highest
dialectness label (\textit{Most}) and the third assigns at least \textit{Little}. These
two values therefore mark natural boundaries between minimal dialectal signal
($a_i < 0.11$), strong dialectal signal ($a_i > 0.77$), and an intermediate range
characterized by mixed or graded annotator judgments.

\begin{figure}[t]
\centering
\includegraphics[width=\linewidth]{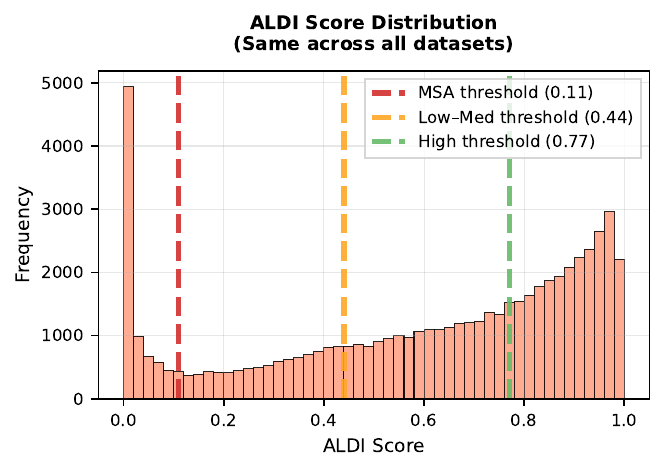}
\caption{Distribution of ALDi scores in the dataset. Vertical dashed lines indicate the
thresholds distinguishing MSA text ($a_i < 0.11$), low--medium dialectness
($0.11$--$0.44$), medium dialectness ($0.44$--$0.77$), and highly dialectal text
($a_i > 0.77$).}
\label{fig:aldi-distribution}
\end{figure}

\paragraph{ALDi Score Distribution.}
ALDi scores are concentrated near the MSA ($a_i < 0.11$) and highly dialectal
($a_i > 0.77$) ranges, with fewer samples in the mid-range
($0.11 \le a_i \le 0.77$), as shown in \autoref{fig:aldi-distribution}.

\section{Non-Neighbouring Dialect Regions}
\label{app:non-neighbouring}

When constructing negative samples for the binary dialect classifiers
(\autoref{sec:dataset}), we restrict negatives to sentences geolocated in
non-neighbouring countries. Neighbouring countries are defined as those sharing a land border with the target country, and negative samples are drawn exclusively from
non-bordering countries.

By restricting negative samples to non-bordering countries, we avoid cases where
sentences may be linguistically compatible with the target dialect, resulting in more
reliable negative supervision.
\section{Dataset Diagnostics}
\label{app:diagnostics}

This appendix presents diagnostics of the constructed multi-label dataset, focusing on
how label cardinality varies with the Arabic Level of Dialectness (ALDi) score for the
supervision sources considered in this work.

\subsection{Label Cardinality Across ALDi Ranges}
\begin{figure*}[t]
    \centering
    \includegraphics[width=\textwidth]{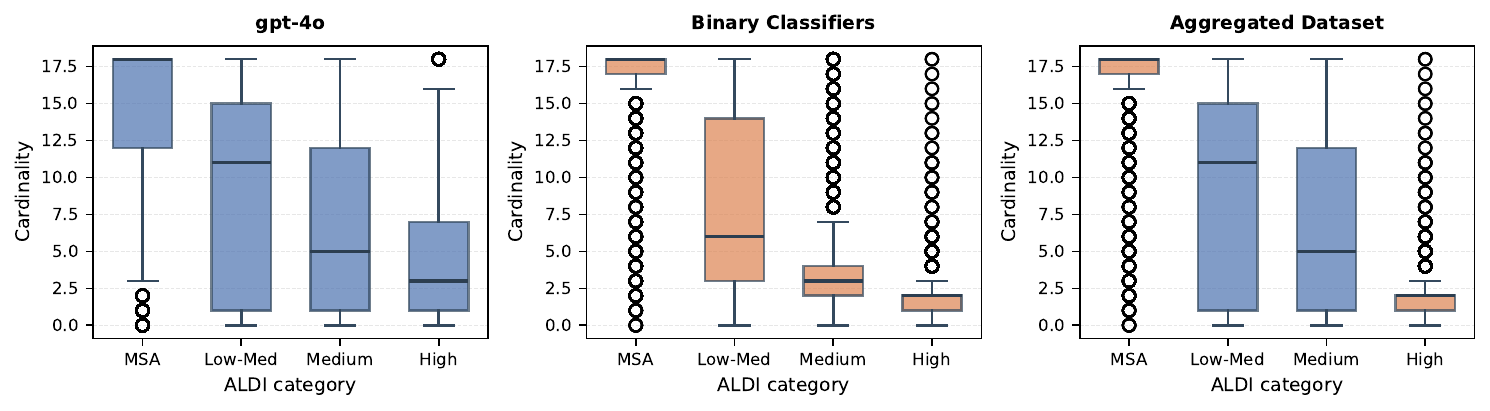}
    \caption{
    Cardinality across ALDi categories for the three labeling methods.
    Due to instabilities, we rely on GPT-4o for
    ambiguous cases and on the binary classifiers when the dialectness is clearly low
    (MSA) or high. This leads to more consistent and linguistically plausible
    multi-label patterns.
    }
    \label{fig:cardinality-ALDi-boxplot}
\end{figure*}

\autoref{fig:cardinality-ALDi-boxplot} shows the distribution of label cardinality
across ALDi categories for each labeling method, as well as for the aggregated dataset.

\paragraph{Binary Dialect Classifiers.}
The binary classifiers exhibit a strong dependence on the ALDi score. For sentences with
low ALDi values (MSA text), they frequently activate many dialect labels, resulting
in high cardinality. For sentences with high ALDi values (strongly dialectal text), they
tend to activate very few labels. In the intermediate ALDi ranges
($0.11 \le a_i \le 0.77$), where explicit negative supervision is absent, classifier
outputs concentrate at either high or low cardinalities, and intermediate label counts
are rarely observed.

\paragraph{GPT-4o.}
GPT-4o displays a smoother relationship between ALDi and label cardinality. While
cardinality generally decreases as ALDi increases, GPT-4o assigns moderate numbers of
labels more frequently in the Low–Med and Medium ALDi ranges. This pattern is consistent
with graded representations of dialectal overlap and contrasts with the polarized
outputs of the binary classifiers in the same ALDi regions.

\paragraph{Aggregation Effects.}
The aggregated dataset (right panel of \autoref{fig:cardinality-ALDi-boxplot})
combines these two behaviors. At low and high ALDi values, where binary classifier
predictions are stable, the aggregated cardinality closely matches their outputs. In
the intermediate ALDi ranges, where binary classifier predictions are concentrated at
extreme cardinalities, the aggregation relies on GPT-4o, resulting in intermediate
label counts and a smoother dependence of cardinality on ALDi.

\subsection{Overall Label Cardinality Distribution}

\autoref{fig:cardinality-distribution} shows the overall distribution of label cardinality
for the supervision sources used in this work, independent of ALDi conditioning.

\section{Prompt Template for GPT-Based Annotation}
\label{appendix:prompt}

\begin{figure}[H]
\centering
\begin{tcolorbox}[colframe=black, colback=gray!5, arc=2mm, boxrule=0.5mm, width=\linewidth]
\small
\textbf{Instruction:} You are a native Arabic speaker and highly qualified linguist with expert-level understanding of regional Arabic dialects.

Given the sentence provided, evaluate its dialectal characteristics independently for each of the following dialects: Iraq, Egypt, Morocco, Libya, UAE, Saudi Arabia, Bahrain, Syria, Lebanon, Oman, Palestine, Algeria, Jordan, Tunisia, Kuwait, Yemen, Sudan, and Qatar.

Return findings in JSON format:
\begin{verbatim}
{
  "Iraq": 0/1,
  "Egypt": 0/1,
  ...
  "Qatar": 0/1
}
Input sentence: {tweet}
\end{verbatim}
\end{tcolorbox}
\caption{GPT-based annotation prompt template.}
\label{fig:prompt-template}
\end{figure}

\section{Curriculum Learning Ordering}
\label{appendix:curriculum motivation}
To obtain a principled ordering for both the cardinality-based curriculum learning and the ALDi-based curriculum learning, we first train a baseline MARBERT model on the aggregated multi-label dataset without any curriculum learning. After fine-tuning, we freeze this baseline model and run inference on the full training set, recording the per-example loss.

\begin{figure}
    \centering
    \includegraphics[width=1\linewidth]{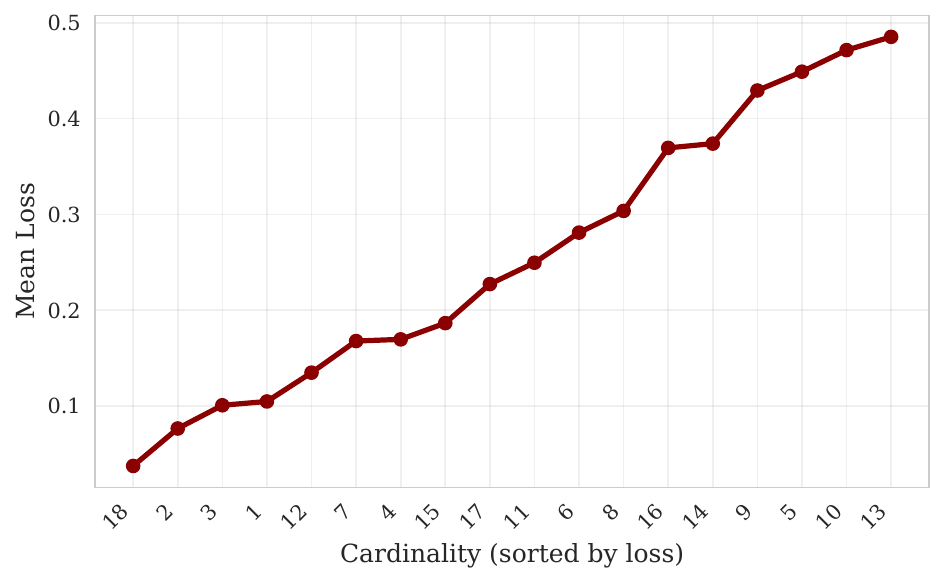}
    \caption{Sorted mean loss per cardinality, to measure the difficulty for the baseline model in predicting different cardinalities.}
    \label{fig:cardinality_loss}
\end{figure}

\begin{figure}
    \centering
    \includegraphics[width=1\linewidth]{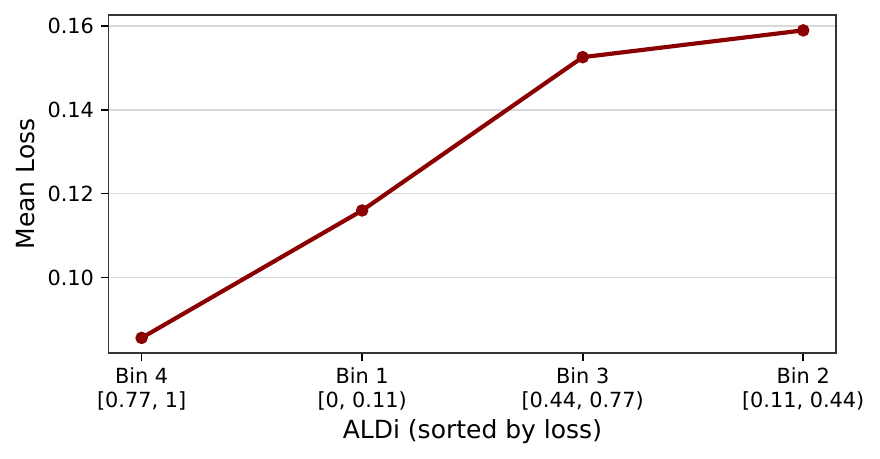}
    \caption{Sorted mean loss per ALDi bin, to measure the difficulty for the baseline model in predicting different cardinalities.}
    \label{fig:ALDi_loss}
\end{figure}

We then aggregate these losses in two ways: (i) by label cardinality, computing the mean loss for each cardinality value, and (ii) by ALDi bin, also computing the mean loss but for each ALDi interval. The resulting mean-loss profiles are visualized in \autoref{fig:cardinality_loss} (mean loss per cardinality) and \autoref{fig:ALDi_loss} (mean loss per ALDi bin).

We treat the mean loss of a bucket as a proxy for its difficulty, and order the stages from lower-loss (easier) buckets to higher-loss (harder) buckets. This difficulty-based ordering directly determines the progression of stages in both the cardinality-based and ALDi-based curriculum learning schedules described in \autoref{sec:multilabel-modeling}.


\section{Model's Predictions on MADAR's Samples}
\label{sec:predictions_contrast}
To analyze the behavior of the three newly introduced models, we study their predictions on 200 sentences from the MADAR-26 corpus \cite{salameh-etal-2018-fine}, for each of the 6 anchor dialects identified by the corpus creators. Additionally, we contrast their predictions to two baseline models: a single-label DI model \textbf{DID-Country}, and the multi-label DI \textbf{NADI2024 shared task's baseline}.

Comparing our three \textsc{LahjatBERT} models to the NADI2024 baseline, it is clear that the baseline is more susceptible to predicting the sentence to be acceptable in countries from different regions than our newly introduced model. For instance, out of the 200 sentences in Rabat's dialect (spoken in Morocco), the baseline model unintuitively predicts that 190 of them are acceptable in Egypt, in comparison to less than 20 of them predicted to be acceptable by the three \textsc{LahjatBERT} models. Hence, assuming that the labels within the top-p (p=90\%) of a single-label DI model as acceptable dialects for the input sentence is not an optimal strategy for multi-label dialect identification.

For the three \textsc{LahjatBERT} models, the Cardinality-based one seems to frequently predict labels from other regions, which explains its lower precision yet higher recall than the two other models. The ALDi-based model seem to be achieving the best compromise between the precision and the recall, as indicated by the fact it achieves the highest overall macro-F1 score.

\begin{table*}[hbpt]
    \centering
    \scriptsize
    \begin{tabular}{p{0.75cm}P{2.5cm}P{2.75cm}P{2.5cm}P{2.5cm}P{2.5cm}}
        \multirow{2}{*}{\textbf{Dialect}} & \multirow{1}{*}{\textbf{DID-Country}} & \multirow{1}{*}{\textbf{NADI2024 Baseline}} & \multirow{2}{*}{\textbf{\textsc{LahjatBERT}}} & \textbf{\textsc{LahjatBERT} +} & \textbf{\textsc{LahjatBERT} +} \\
        & \cite{salameh-etal-2018-fine} & \cite{abdul-mageed-etal-2024-nadi} &  & \textbf{ALDi CL} & \textbf{Cardinality CL} \\
        \midrule
        BEI (200) & \textbf{LB~129}, \underline{SY~32}, \underline{JO~15}, \underline{PS~12}, EG~4, SA~3, IQ~2, OM~2, MA~1 & \textbf{LB~200}, \underline{SY~200}, \underline{JO~199}, \underline{PS~199}, EG~180, IQ~174, TN~160, SD~127, BH~86, MA~78, KW~68, DZ~57, SA~56, QA~53, LY~41, OM~36, YE~25, AE~19 & \underline{SY~165}, \textbf{LB~149}, \underline{PS~131}, \underline{JO~121}, IQ~36, SA~35, AE~34, KW~32, BH~31, OM~31, QA~29, YE~27, EG~23, LY~18, SD~17, TN~10, DZ~9, MA~9 & \underline{SY~174}, \textbf{LB~146}, \underline{PS~135}, \underline{JO~129}, OM~35, SA~33, AE~33, IQ~31, KW~31, BH~30, QA~30, YE~26, EG~25, SD~19, LY~18, TN~9, DZ~8, MA~7 & \underline{SY~188}, \textbf{LB~184}, \underline{PS~181}, \underline{JO~173}, SA~45, AE~44, BH~42, OM~42, KW~41, QA~41, YE~41, IQ~35, EG~31, SD~23, LY~22, MA~9, TN~9, DZ~7 \\
        \midrule
        CAI (200) & \textbf{EG~147}, SA~12, SY~12, \underline{SD~9}, JO~7, DZ~5, LY~4, TN~4, MA~3, YE~3, IQ~2, PS~2, QA~2, KW~1, LB~1, BH~1, OM~1, AE~1 & \textbf{EG~200}, YE~193, SA~192, \underline{SD~192}, PS~188, JO~187, LY~182, LB~178, SY~167, MA~159, OM~142, TN~130, IQ~73, KW~54, QA~51, AE~40, DZ~20 & \textbf{EG~190}, PS~131, \underline{SD~115}, LB~54, SA~54, JO~52, SY~52, LY~51, KW~46, IQ~44, BH~44, QA~44, AE~43, OM~42, YE~42, TN~14, MA~12, DZ~10 & \textbf{EG~188}, PS~152, \underline{SD~132}, LY~49, SA~49, JO~46, LB~45, IQ~44, SY~44, KW~43, BH~41, OM~41, QA~41, AE~41, YE~41, TN~11, MA~9, DZ~8 & PS~193, \textbf{EG~189}, \underline{SD~165}, SA~135, LB~103, JO~99, SY~93, QA~83, OM~81, AE~80, BH~79, YE~79, KW~74, LY~73, IQ~60, TN~14, MA~12, DZ~10 \\
        \midrule
        DOH (200) & \textbf{QA~144}, \underline{SA~17}, IQ~7, JO~5, SY~5, \underline{OM~4}, MA~3, SD~3, TN~3, PS~3, LY~2, EG~2, YE~2 & \underline{SA~200}, \textbf{QA~199}, \underline{OM~198}, LY~194, JO~191, SY~181, IQ~163, SD~156, YE~137, \underline{AE~122}, MA~112, \underline{KW~86}, LB~80, TN~77, PS~76, \underline{BH~64}, EG~59, DZ~14 & \underline{SA~173}, \underline{KW~154}, \underline{BH~142}, \underline{AE~136}, \textbf{QA~133}, \underline{OM~128}, IQ~115, YE~111, PS~93, JO~91, SY~91, LB~89, SD~55, LY~54, EG~52, MA~24, TN~24, DZ~21 & \underline{SA~179}, \underline{KW~165}, \underline{BH~151}, \textbf{QA~147}, \underline{AE~142}, \underline{OM~134}, YE~117, IQ~111, PS~105, LB~102, JO~100, SY~100, EG~53, SD~49, LY~46, MA~17, TN~17, DZ~14 & \underline{SA~197}, \underline{BH~194}, \textbf{QA~193}, \underline{KW~190}, \underline{AE~190}, \underline{OM~175}, YE~149, IQ~132, PS~115, JO~111, LB~96, SY~96, SD~53, EG~50, LY~46, MA~14, TN~13, DZ~11 \\
        \midrule
        RAB (200) & \textbf{MA~176}, \underline{DZ~6}, \underline{TN~4}, JO~3, \underline{LY~3}, SA~3, PS~2, OM~1, SY~1, QA~1 & \textbf{MA~198}, \underline{DZ~196}, \underline{LY~194}, EG~190, QA~188, LB~180, AE~165, SA~159, SY~154, SD~151, \underline{TN~151}, PS~147, KW~143, BH~125, IQ~70, JO~41, OM~32, YE~11 & \underline{DZ~172}, \textbf{MA~171}, \underline{TN~84}, \underline{LY~27}, EG~16, PS~15, IQ~13, SY~13, JO~11, LB~11, SA~11, OM~10, AE~10, KW~9, BH~8, QA~8, YE~8, SD~7 & \underline{DZ~181}, \textbf{MA~175}, \underline{TN~68}, \underline{LY~23}, PS~14, EG~10, IQ~8, SY~8, JO~7, SA~7, LB~6, AE~6, QA~6, KW~5, BH~5, OM~5, YE~4, SD~1 & \textbf{MA~178}, \underline{DZ~174}, \underline{TN~130}, \underline{LY~52}, PS~32, SY~30, JO~29, LB~27, AE~25, KW~24, OM~24, SA~24, QA~24, YE~24, IQ~21, BH~20, EG~19, SD~16 \\
        \midrule
        TUN (200) & \textbf{TN~167}, \underline{DZ~6}, SA~5, JO~4, IQ~3, OM~3, PS~3, \underline{LY~2}, \underline{MA~2}, SY~2, QA~2, YE~1 & \underline{LY~198}, \textbf{TN~197}, LB~192, IQ~191, \underline{DZ~190}, EG~189, \underline{MA~183}, OM~181, SD~180, QA~160, YE~150, KW~104, PS~99, AE~60, SY~42, SA~40, JO~28, BH~14 & \textbf{TN~159}, \underline{LY~149}, \underline{DZ~143}, \underline{MA~94}, IQ~24, SY~24, PS~24, LB~23, JO~21, EG~19, KW~17, SA~17, AE~16, YE~15, BH~14, OM~14, SD~14, QA~14 & \textbf{TN~163}, \underline{DZ~155}, \underline{LY~153}, \underline{MA~57}, PS~28, IQ~25, JO~23, LB~23, SY~23, EG~21, KW~20, SA~20, AE~20, OM~19, YE~18, BH~17, QA~16, SD~15 & \underline{LY~170}, \textbf{TN~162}, \underline{DZ~159}, \underline{MA~103}, PS~48, SY~44, JO~42, LB~41, IQ~35, SA~33, KW~32, OM~31, AE~31, QA~30, YE~30, BH~29, EG~28, SD~21 \\
        \bottomrule \addlinespace[0.5em]
        MSA (200) & \textbf{OM~162}, \textbf{SA~162}, \textbf{SD~152}, \textbf{LY~151}, \textbf{SY~151}, \textbf{DZ~150}, \textbf{IQ~149}, \textbf{JO~149}, \textbf{EG~149}, \textbf{PS~148}, \textbf{KW~147}, \textbf{LB~147}, \textbf{BH~147}, \textbf{MA~147}, \textbf{QA~147}, \textbf{TN~147}, \textbf{AE~147}, \textbf{YE~147} & \textbf{OM~200}, \textbf{SA~200}, \textbf{SD~200}, \textbf{IQ~195}, \textbf{YE~189}, \textbf{JO~165}, \textbf{AE~151}, \textbf{QA~145}, \textbf{LY~144}, \textbf{EG~143}, \textbf{DZ~111}, \textbf{KW~100}, \textbf{BH~91}, \textbf{TN~75}, \textbf{MA~57}, \textbf{SY~56}, \textbf{PS~30}, \textbf{LB~22} & \textbf{IQ~196}, \textbf{JO~195}, \textbf{LB~195}, \textbf{PS~195}, \textbf{SA~194}, \textbf{SY~194}, \textbf{AE~194}, \textbf{KW~193}, \textbf{LY~193}, \textbf{EG~193}, \textbf{OM~192}, \textbf{YE~192}, \textbf{BH~191}, \textbf{SD~191}, \textbf{QA~190}, \textbf{TN~188}, \textbf{MA~187}, \textbf{DZ~183} & \textbf{IQ~198}, \textbf{EG~198}, \textbf{LY~197}, \textbf{PS~197}, \textbf{SA~197}, \textbf{JO~196}, \textbf{KW~196}, \textbf{LB~196}, \textbf{SD~196}, \textbf{SY~196}, \textbf{AE~196}, \textbf{YE~196}, \textbf{BH~195}, \textbf{OM~195}, \textbf{QA~195}, \textbf{MA~189}, \textbf{TN~189}, \textbf{DZ~188} & \textbf{IQ~200}, \textbf{JO~200}, \textbf{KW~200}, \textbf{BH~200}, \textbf{OM~200}, \textbf{PS~200}, \textbf{QA~200}, \textbf{SA~200}, \textbf{AE~200}, \textbf{YE~200}, \textbf{LB~199}, \textbf{SY~199}, \textbf{LY~197}, \textbf{SD~197}, \textbf{EG~196}, \textbf{MA~190}, \textbf{TN~190}, \textbf{DZ~188} \\
    \bottomrule
    \end{tabular}
    \caption{The number of times a country label is predicted by the model for the 200 sentences of the six anchor dialects of the MADAR CORPUS-26 corpus \cite{salameh-etal-2018-fine}: Beirut~(BEI), Cairo~(CAI), Doha~(DOH), Rabat~(RAB), Tunis~(TUN), and MSA. Each dialect's representative country is in \textbf{bold}, and the country's same-region countries are \underline{underlined}, following the regional grouping of \citet{baimukan-etal-2022-hierarchical}. \textbf{Note \#1:} the DID-country model is a single-label DI model trained on the MADAR corpus. Its province-level predictions are mapped into country-level ones, with MSA predictions mapped to all the 18 country-level dialects we consider. \textbf{Note \#2:} Country abbreviations - \textbf{AE}: UAE, \textbf{BH}: Bahrain, \textbf{DZ}: Algeria, \textbf{EG}: Egypt, \textbf{IQ}: Iraq, \textbf{JO}: Jordan, \textbf{KW}: Kuwait, \textbf{LB}: Lebanon, \textbf{LY}: Libya, \textbf{MA}: Morocco, \textbf{OM}: Oman, \textbf{PS}: Palestine, \textbf{QA}: Qatar, \textbf{SA}: Saudi Arabia, \textbf{SD}: Sudan, \textbf{SY}: Syria, \textbf{TN}: Tunisia, \textbf{YE}: Yemen.}
    \label{tab:TODO}
\end{table*}
\end{document}